\documentclass[runningheads]{llncs}
\usepackage{graphicx}
\usepackage[table]{xcolor}
\usepackage{url}
\usepackage{siunitx}

\usepackage{color}

\usepackage{subcaption}

\begin{document}
%


\title{A Planning Ontology to Represent and Exploit Planning Knowledge for Performance Efficiency}

\titlerunning{Planning Ontology}


%


\author{Bharath Muppasani\inst{1} \and
Vishal Pallagani\inst{1} \and
Biplav Srivastava\inst{1} \and
Raghava Mutharaju\inst{2} \and
Michael N. Huhns\inst{1} \and
Vignesh Narayanan\inst{1}
}
\authorrunning{B. Muppasani et al.}

%
\institute{University of South Carolina, USA \and IIIT-Delhi, India \\
\email{bharath@email.sc.edu},
\email{vishalp@mailbox.sc.edu},
\email{biplav.s@sc.edu},
\email{raghava.mutharaju@iiitd.ac.in},
\email{huhns@sc.edu},
\email{vignar@sc.edu}
}
\maketitle              
\begin{abstract}

Ontologies are known for their ability to organize rich metadata, support the identification of novel insights via semantic queries, and promote reuse. In this paper, we consider the problem of automated planning, where the objective is to find a sequence of actions that will move an agent from an initial state of the world to a desired goal state. We hypothesize that given a large number of available planners and diverse planning domains; they carry essential information that can be leveraged to identify suitable planners and improve their performance for a domain. We use data on planning domains and planners from the International Planning Competition (IPC) to construct a planning ontology and demonstrate via experiments in two use cases that the ontology can lead to the selection of promising planners and improving their performance using macros - a form of action ordering constraints extracted from planning ontology. We also make the planning ontology and associated resources available to the community to promote further research. 

     \\
    
    \textbf{Resource Type:} Ontology, Knowledge Graph \\
    \textbf{Licence:} Creative Commons Attribution 4.0 License \\
    \textbf{URL:} \url{https://github.com/BharathMuppasani/AI-Planning-Ontology}

\end{abstract}
\section{Introduction}

Automated planning, where the objective is to find a sequence of actions that will transition an agent from the initial state of the world to a desired goal state, is an active sub-field of Artificial Intelligence (AI). The ability to generate plans and make decisions in complex domains, such as robotics, logistics, and manufacturing, has led to significant progress in the automation of planning. Currently, there are numerous planning domains, planners, search algorithms, and associated heuristics in the field of automated planning. Each planner, in conjunction with a search algorithm and heuristic, generates plans with varying degrees of quality, cost, and optimality. The empirical results available for various planning problems, ranked by planner performance and the heuristics used as available in International Planning Competition (IPC), can provide valuable information to identify various tunable parameters to improve planner performance. Traditionally, improving planner performance involves manually curating potential combinations to identify the optimal planner configuration. However, there has been limited effort to model the available information in a structured knowledge representation, such as an ontology, to facilitate efficient reasoning and enhance planner performance.

To address the challenge of representing planning problems and associated information in a structured manner, we propose an ontology for AI planning. An ontology formally represents concepts and their relationships \cite{guarino2009ontology}, which enables systematic analysis of planning domains and planners. The proposed ontology captures the features of a domain and the capabilities of planners, facilitating reasoning with existing planning problems, identifying similarities, and suggesting different planner configurations. Planning ontology can also be a useful resource for the creation of new planners as it captures essential information about planning domains and planners, which can be leveraged to design more efficient planning algorithms. Furthermore, ontology can promote knowledge sharing and collaboration within the planning community.
  
In the field of planning, several attempts have been made to create ontologies to enhance the understanding of planners' capabilities. For instance, Plan-Taxonomy \cite{plan-taxonomy} introduced a taxonomy that aimed to explain the functionality of planners. Additionally, authors in \cite{gil2000planet} present a comprehensive ontology called PLANET, which represents plans in real-world domains and can be leveraged to construct new applications. Nonetheless, the reusability of PLANET is limited as it is not open-sourced. Consequently, researchers face difficulty in extending or replicating the ontology.



This paper outlines our methodology for constructing an ontology to represent AI planning domains, leveraging information obtained from the IPC. In our current work, we extended our initial research \cite{planning-ontology}. Specifically, we have enhanced the ontology to more accurately depict the various concepts within the planning domain. Furthermore, we include additional use cases of our ontology and provide experimental evaluations to support our findings further. Building a planning ontology using data from IPC offers several benefits, such as comprehensive coverage of planning domains, a rich source for various benchmark evaluation metrics, and documentation for planners. However, the ontology is not limited to the PDDL representation or  domains in IPC and can easily  be extended to any. 
Our contributions are at the intersection of ontologies and AI planning and can be summarized as follows.
\begin{itemize}
    \item \textbf{Planning Ontology}: We developed an ontology for AI planning that can be used to represent and organize knowledge related to planning problems. We designed the competency questions to ensure that our ontology provides a structured way to capture the relationships between different planning concepts, enabling more efficient and effective knowledge sharing and reuse.

    \item \textbf{Usecase 1: Identifying Most Promising Planner for Performance}: We demonstrate the ontology's usage for identifying the best-performing planner for a specific planning domain using data from IPC-2011.


    \item \textbf{Usecase 2: Macro Selection for Improving Planner Performance}: We demonstrate the usage of ontology to extract domain-specific macros - which are action orderings and show that they can improve planner performance drastically.

\end{itemize}

In the remainder of the paper, we start with preliminaries about on automated planning and IPC. We then give an overview of the existing literature on ontologies for planning. Following this, we present a detailed description of the ontology construction process and its usage. We then discuss the proposed planning ontology and conclude with future research directions.
\section{Preliminaries}
In this section, we describe the necessary background for automated planning and the significance of the International Planning Competition.

\begin{figure}[!b]
    \centering
    \includegraphics[scale=0.37]{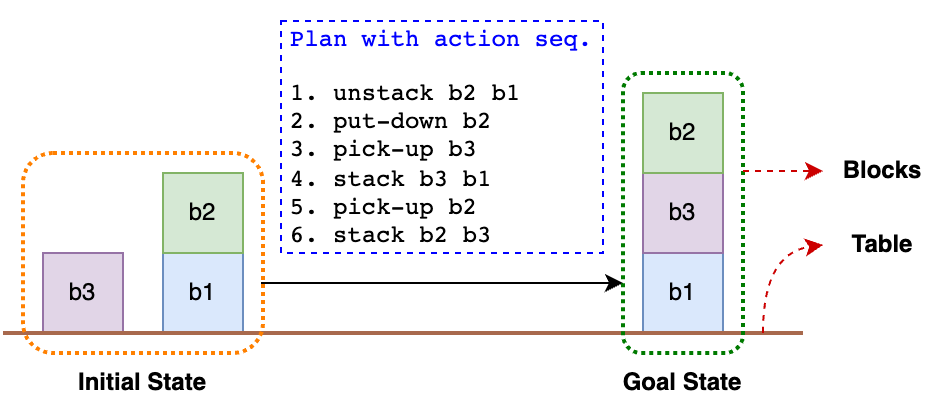}
    \caption{Demonstration of automated planning problem with blocksworld domain example}
    \label{fig:planning_bw}
\end{figure}

\subsection{Automated Planning}

Automated planning, also known as AI planning, is the process of finding a sequence of actions that will transform an initial state of the world into a desired goal state \cite{ghallab2004automated}. It involves constructing a plan or a sequence of actions that will achieve a specified objective while respecting any constraints or limitations that may be present. Formally, automated planning can be defined as a tuple $(S, A, T, I, G)$, where:
\begin{itemize}
    \item $S$ is the set of possible states of the world
    \item $A$ is the set of possible actions that can be taken
    \item $T$ is the transition function that describes the effects of taking an action on the current state of the world
    \item $I$ is the initial state of the world
    \item $G$ is the desired goal state
\end{itemize}
Using this notation, the problem of automated planning can be framed as finding a sequence of actions $\prec a_1, a_2, ..., a_k\succ$ that will transform the initial state $I$ into the goal state $G$, while respecting any constraints or limitations on the actions. 
 A problem is defined in terms of a domain and a problem instance. The domain defines the possible actions that can be taken and the effects of each action, while the problem instance specifies the initial state of the world and the desired goal state. 
Various techniques can be used to solve the planning problem, such as search algorithms, constraint-based reasoning, and optimization methods. These techniques involve exploring the space of possible plans and selecting the one that satisfies the objective and any constraints. Figure \ref{fig:planning_bw} illustrates an automated planning scenario for the blocksworld domain, where an initial state can be transformed into a goal state by executing a sequence of actions.

\noindent \textbf{Attributes modeled about a domain.}
 \begin{enumerate}
     \item \textbf{Requirements:} A list of requirements that the planner must satisfy to solve the given domain, e.g., \emph{typing} in blocksworld with types.
     \item \textbf{Predicates:} Define world properties, e.g., \verb|(on b1 b2)| in blocksworld.
     \item \textbf{Actions:} Units of change with preconditions and effects, e.g., \verb|unstack b2 b1| in blocksworld.
     \item \textbf{Preconditions:} Conditions for action execution, e.g., \verb|(on b1 b2)| for \\ \verb|unstack b2 b1|.
     \item \textbf{Effects:} Post-action world changes, e.g., \verb|(not (on b1 b2))| after \\ \verb|unstack b2 b1|.
     \item \textbf{Constants:} Fixed values, e.g., \emph{table} in blocksworld.
     \item \textbf{Types:} Classifications based on attributes, e.g., \\ \verb|(on ?x - block ?y - block)| in typed blocksworld.
 \end{enumerate}

\noindent \textbf{Attributes modeled about a problem instance from a domain.}
\begin{enumerate}
    \item \textbf{Name:} The name of the planning problem.
    \item \textbf{Domain:} The name of the planning domain that the problem belongs to.
    \item \textbf{Objects:} A list of objects that are present in the planning problem. Objects are typically defined in terms of their type and name. In the example shown in Figure \ref{fig:planning_bw}, objects are b1, b2, and b3.
    \item \textbf{Initial State:} A description of the initial state of the world, including the values of all relevant predicates. Figure \ref{fig:planning_bw} represents an example initial state.
    \item \textbf{Goal State:} A description of the desired goal state of the world, including the values of all relevant predicates. Figure \ref{fig:planning_bw} represents an example goal state.
\end{enumerate}

\subsection{International Planning Competition (IPC)}


IPC is pivotal for evaluating and contrasting planning systems. Introducing new planners and benchmarks, it promotes innovative planning methodologies and reflects the field's evolving challenges. The competition has multiple tracks, such as classical and probabilistic planning, with benchmarks assessing plan quality, length, and run time. IPC results offer a glimpse into the latest in planning, highlighting system pros and cons. The benchmarks from IPC are ideal for crafting a planning-related ontology, encapsulating the domain's breadth and planners' challenges.

\section{Related Work}

The use of ontology-based knowledge representation and reasoning has been extensively studied in various domains, including automated planning. This section focuses on the applications of ontology-based knowledge representation and reasoning in the context of planning and related domains.
In \cite{valente1999building}, an ontology is constructed for the Joint Forces Air Component Commander (JFACC) to represent knowledge from the air campaign domain. The ontology is modularized to facilitate data organization and maintenance, but its applicability is domain-specific, unlike our approach. In \cite{vzakova2010automating}, the authors automate the knowledge discovery workflow using ontology and AI planning, creating a Knowledge Discovery (KD) ontology to represent the KD domain and converting its variables to a Planning Domain Definition Language (PDDL) format to obtain the PDDL domain. The ontology's objects represent initial and goal states, forming the KD task, which represents a specific problem. The authors use the Fast-Forward (FF) planning system to generate the required plans.

In a survey of ontology-based knowledge representation and reasoning in the planning domain, \cite{gayathri2018ontology} suggests that knowledge reasoning approaches can draw new conclusions in non-deterministic contexts and assist with dynamic planning. In \cite{gil2000planet}, a reusable ontology, PLANET, is proposed for representing plans. PLANET includes representations for planning problem context, goal specification, plan, plan task, and plan task description. However, PLANET does not include representations for some entities commonly associated with planning domains, such as resources and time. Our planning ontology draws inspiration from PLANET and appends more metadata for planner improvement.
In \cite{babli2019extending}, a domain-independent approach is presented that advances the state of the art by augmenting the knowledge of a planning task with pertinent goal opportunities. The authors demonstrate that incorporating knowledge obtained from an ontology can aid in producing better-valued plans, highlighting the potential for planner enhancement using more tuning parameters, which are captured in our planning ontology. The CARESSES ontology \cite{khaliq2018culturally} is another significant development in planning-oriented ontologies, focusing on cultural competence in socially assistive robots for elderly care. Our work incorporates aspects from this ontology, specifically the concepts of \texttt{Action} and \texttt{Parameter}. In general, these studies demonstrate the potential of ontology-based knowledge representation and reasoning in the planning domain, including applications such as representing plans, aiding in air campaign planning, automating knowledge discovery workflows, and developing context-aware planning services.

\begin{table}[b]
\centering
\caption{Concepts reused from various ontologies}
\begin{tabular}{ll}
\hline
\textbf{Concept} & \textbf{Ontology} \\ \hline
Action           & http://caressesrobot.org/ontology \cite{khaliq2018culturally} \\ 
Parameter        & http://caressesrobot.org/ontology \cite{khaliq2018culturally} \\ 
Plan             & http://www.ontologydesignpatterns.org/ont/dul/DUL.owl \cite{mascardi2008comparison} \\
State            & http://purl.org/vocab/lifecycle/schema \\

\hline
\end{tabular}
\label{tab:ontology-concepts}
\end{table}


\section{Planning Ontology}

\begin{figure}[!t]
    \centering
    \includegraphics[scale=0.19]{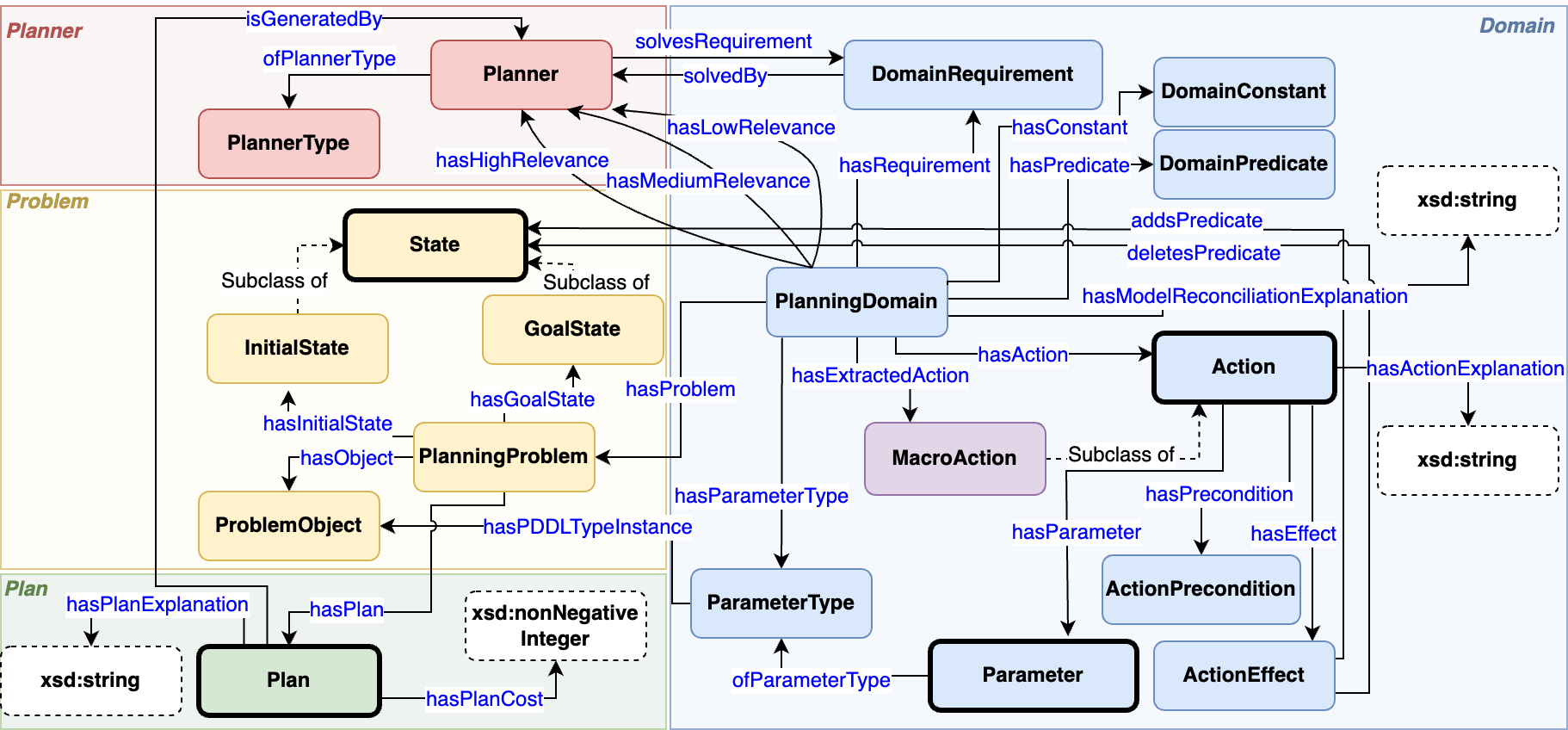}
    \caption{An illustrative overview of the planning ontology, segmented into categories that encapsulate the core concepts of automated planning: domain, problem, plan, and planner performance. Each category is distinctly represented by colored rectangles. Classes with thick outlines denote concepts that have been adapted or reused from existing ontologies.}
    \label{fig:ontology}
\end{figure}


This section covers the construction of planning ontology to capture the essential details of automated planning. We will discuss the considerations, challenges, benefits, and limitations of using ontologies for automated planning, to provide a better understanding of how they can improve the efficiency and effectiveness of automated planning systems.

\subsection{Competency Questions}

Competency questions for an ontology are focused on the needs of the users who will be querying the ontology. These questions are designed to help users explore and understand the concepts and relationships within the ontology, and to find the information they need within the associated knowledge base. By answering these questions, the ontology can be better scoped and tailored to meet the needs of its users. 

We designed the following competency questions to model an Ontology to represent the general aspects of Automated Planning.

\begin{itemize}
    \item C1: What are the different types of planners used in automated planning?
    \item C2: What is the relevance of planners in a given problem domain?
    \item C3: What are the available actions for a given domain?
    \item C4: What problems in a domain satisfy a given condition?
    \item C5: What are all the requirements a given domain has?
    \item C6: What is the cost associated with generating a plan for a given problem?
    \item C7: How many parameters does a specific action have?
    \item C8: What planning type does a specific planner belong to?
    \item C9: What requirements does a given planner support?
    \item C10: What are the different parameter \verb|types| present in a domain?
\end{itemize}

\subsection{Design}
An ontology is a formal and explicit representation of concepts, entities, and their relationships in a particular domain. In this case, ontology is concerned with the domain of automated planning, which refers to the process of generating a sequence of actions to achieve a particular goal within a given set of constraints. The ontology aims to provide a structured framework for organizing and integrating knowledge about this domain, which can be useful in various applications, such as designing planning algorithms, extracting best-performing planners given a domain, or learning domain-specific macros.

Figure \ref{fig:ontology} shows an ontology that aims to encompass the various concepts of automated planning separated into categories of \verb|Domain|, \verb|Problem|, \verb|Plan|, and \verb|Planner|. The ontology for automated planning is composed of 19 distinct classes and 25 object properties. These classes and properties are designed to represent the various elements of the automated planning domain and its associated problems. In the design of our ontology, all axioms are formulated using Description Logic \cite{KSH14:DLintro}, providing a formal and expressive framework for representing and reasoning about the concepts and relationships within our domain.

\subsubsection{Domain}
The Domain category in our ontology comprises the characteristics of the AI planning domain through several classes. These include \texttt{PlanningDomain} - \texttt{DomainRequirement}, detailing domain modeling; \texttt{ParameterType}, defining parameter varieties in a typed domain; \texttt{DomainPredicate}, encompassing applicable predicates; \texttt{DomainConstant}, representing invariant constants; and \texttt{Action}, for domain operations. \texttt{Action} class is further linked with \texttt{ActionPrecondition}, \texttt{ActionEffect}, and \texttt{Parameter}. This structured approach aids applications like algorithm design, planner optimization, and macro learning in domain-specific contexts.

The \texttt{PlanningDomain} conceptualization is articulated through axioms to represent fundamental elements of planning scenarios. Axiom~\ref{ax: domain1} signifies that every planning domain entails certain actions. Actions are fundamental to planning as they represent the steps or decisions that can be taken to transform a state within the domain. Predicates are essential for defining the states within a planning domain. Axiom~\ref{ax: domain2} ensures that each domain includes predicates to represent these states, facilitating the definition of preconditions and effects of actions. Axiom~\ref{ax: domain3} states that every planning domain possesses certain defined requirements. Requirements in AI Planning are necessary to define various types of domain modeling, such as conditional effects and numeric fluents. Such specifications are not only essential for characterizing the domain but also serve as a criterion to assess whether a planner is compatible with and can support these specific domain modeling features.

\begin{equation}
\texttt{PlanningDomain} \sqsubseteq \exists\texttt{hasAction}.\texttt{Action}
\label{ax: domain1}
\end{equation}
\begin{equation}
\texttt{PlanningDomain} \sqsubseteq \exists\texttt{hasPredicate}.\texttt{DomainPredicate}
\label{ax: domain2}
\end{equation}
\begin{equation}
\texttt{PlanningDomain} \sqsubseteq \exists\texttt{hasRequirement}.\texttt{DomainRequirement}
\label{ax: domain3}
\end{equation}

The \texttt{Action} class is characterized by its effects, a fundamental aspect of planning. Axiom~\ref{ax: action1} addresses the transformative nature of actions in a planning domain. Understanding the effects of actions is essential for planning algorithms to predict and evaluate the outcomes of different action sequences.
\begin{equation}
\texttt{Action} \sqsubseteq \exists\texttt{hasEffect}.\texttt{ActionEffect}
\label{ax: action1}
\end{equation}

Axioms~\ref{ax: action2} and \ref{ax: action3} capture the dynamics of how actions can add or delete predicates in a state, emphasizing the mutable nature of states within the planning domain. This depiction is essential for accurately modeling the consequences and feasibility of actions in AI Planning.
\begin{equation}
\texttt{ActionEffect} \sqsubseteq \exists\texttt{addsPredicate}.\texttt{State}
\label{ax: action2}
\end{equation}
\begin{equation}
\texttt{ActionEffect} \sqsubseteq \exists\texttt{deletesPredicate}.\texttt{State}
\label{ax: action3}
\end{equation}

\subsubsection{Problem}
The Problem category of the ontology includes classes that represent specific problems within a given domain. These classes are designed to capture the details of a particular problem, such as the \verb|Objects| defined in the problem, which is an instance of different \emph{types} defined in the planning domain, the \verb|Initial State| of the problem, and the \verb|Goal State| which are a subclass of the parent class \verb|State| which is a state description of the given domain. 

The axioms defined for \texttt{PlanningProblem} conceptualized the key aspects of a planning problem. Axiom~\ref{ax: problem1} indicates that each planning problem is defined with a specific \texttt{GoalState}, which is the desired outcome or objective of the problem. Axiom~\ref{ax: problem2} asserts that each planning problem also has a defined \texttt{InitialState}, which provides the starting conditions and context for the planning process. Lastly, Axiom~\ref{ax: problem3} identifies the \texttt{Objects} present within a planning problem, denoting the various entities that are subject to manipulation or consideration during the course of planning. Finally, the axiom~\ref{ax: problem4} underscores that every planning problem includes a potential plan or series of actions that lead to the goal state.

\begin{equation}
\texttt{PlanningProblem} \sqsubseteq =1 \texttt{hasGoalState}.\texttt{GoalState}
\label{ax: problem1}
\end{equation}
\begin{equation}
\texttt{PlanningProblem} \sqsubseteq =1 \texttt{hasInitialState}.\texttt{InitialState}
\label{ax: problem2}
\end{equation}
\begin{equation}
\texttt{PlanningProblem} \sqsubseteq \exists\texttt{hasObject}.\texttt{ProblemObject}
\label{ax: problem3}
\end{equation}
\begin{equation}
\texttt{PlanningProblem} \sqsubseteq \exists\texttt{hasPlan}.\texttt{Plan}
\label{ax: problem4}
\end{equation}

\subsubsection{Plan}
The Plan category of the ontology includes classes that represent the sequence of actions that must be taken to solve a given problem. The \verb|Plan| class captures the knowledge about the plans that planners generate for specific problems. The plan cost for each plan is a data property (non-negative integer) of the \verb|Plan| class. This enables planners to be compared based on the quality of the plans they generate and the cost of those plans.

The axioms defined for the \texttt{Plan} category outline the essential features of plans in the AI planning process. Axiom~\ref{ax: plan1} mandates that each plan must have an associated plan cost, precisely quantified as a non-negative integer. This is crucial for evaluating and comparing the efficiency of different plans. Axiom~\ref{ax: plan2} establishes that every plan is generated by some planner, connecting each plan to its generator and allowing for an understanding of the planning process and the assessment of various planners.

\begin{equation}
\texttt{Plan} \sqsubseteq =1 \texttt{hasPlanCost}. \texttt{xsd:nonNegativeInteger}
\label{ax: plan1}
\end{equation}
\begin{equation}
\texttt{Plan} \sqsubseteq \exists\texttt{isGeneratedBy}.\texttt{Planner}
\label{ax: plan2}
\end{equation}

\subsubsection{Planner}
The Planner category of the ontology includes classes that capture the details of the planner, planner type, and the planner performance from previous IPCs. Specifically, \verb|Planning Domain| relevance to a \verb|Planner| is classified based on the percentage of problems they have successfully solved, which is then categorized into three levels of relevance to the planner: \textit{low}, \textit{medium}, and \textit{high}. By incorporating this information into the ontology, planners can be evaluated based on their performance in different planning domains, and more informed decisions can be made. In addition, this information can be used to guide the development of new planners and to evaluate their performance against established benchmarks.

The axioms defined for the \texttt{Planner} category provide a foundation for understanding and assessing the capabilities of planners in the AI planning domain. Axiom~\ref{ax: planner1} classifies planners into different types based on their characteristics or strategies, enabling a nuanced understanding of various planning approaches. Axiom~\ref{ax: planner2} links planners with the specific domain requirements they can solve, highlighting their applicability in different planning scenarios.

\begin{equation}
\texttt{Planner} \sqsubseteq \exists\texttt{ofPlannerType}.\texttt{PlannerType}
\label{ax: planner1}
\end{equation}
\begin{equation}
\texttt{Planner} \sqsubseteq \exists\texttt{solvesRequirement}.\texttt{DomainRequirement}
\label{ax: planner2}
\end{equation}


\subsection{Accessing Planning Ontology}
We have taken various measures to ensure that our planning ontology follows the FAIR principles \cite{wilkinson2016fair} of being Findable, Accessible, Interoperable, and Reusable. To assist users in exploring and utilizing our ontology, we have made it accessible through a persistent URL\footnote[1]{\label{purl}PURL - \url{https://purl.org/ai4s/ontology/planning}} and our GitHub repository\footnote[2]{\label{footnote: repo}\url{https://github.com/BharathMuppasani/AI-Planning-Ontology}}. Our repository contains ontology model files, mapping scripts, and utility scripts that extract information from PDDL domains and problems into intermediary JSON format and add the extracted data as triples using our model ontology, creating a knowledge graph. We provide sample SPARQL queries that address the ontology's competency questions mentioned earlier. Moreover, our ontology documentation, which is accessible through the GitHub repository, provides a comprehensive overview of the ontology's structure, concepts, and relations, including ontology visualization. This documentation serves as a detailed guide for users to comprehend the ontology's applications in the automated planning domain. We also provide the scripts and results from the ontology evaluation, which are presented as use cases of our ontology in later sections, in our repository, along with accompanying documentation.
Furthermore, our commitment includes a proactive approach to constantly updating and refining the ontology. This involves periodic updates and community-driven modifications, ensuring its continuous alignment with evolving standards and practices in the field of automated planning.

\begin{figure}[t]
    \centering
    \includegraphics[scale=0.19]{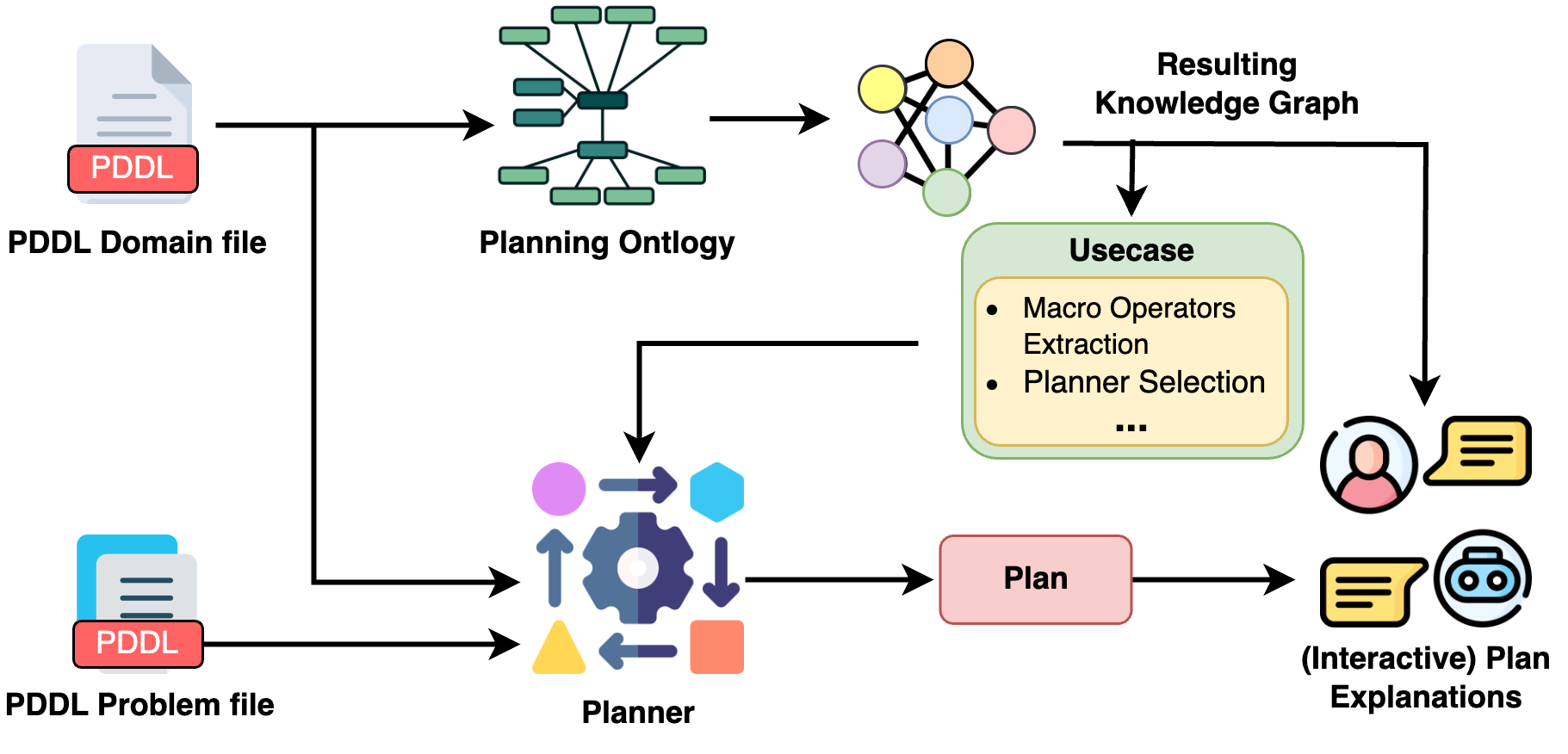}
    \caption{Workflow diagram illustrating the integration of the Planning Ontology with AI Planning system, which supports use cases like Macro Operators Extraction and Planner Selection, ultimately enabling the interactive plan explanations with the help of generated Knowledge graph.}
    \label{fig:use}
\end{figure}

\section{Usage of Planning Ontology}
In the following section, we show the evaluation of a few competency questions and discuss two use cases of our planning ontology.

\begin{figure}[t]
    \centering
    \includegraphics[scale=0.20]{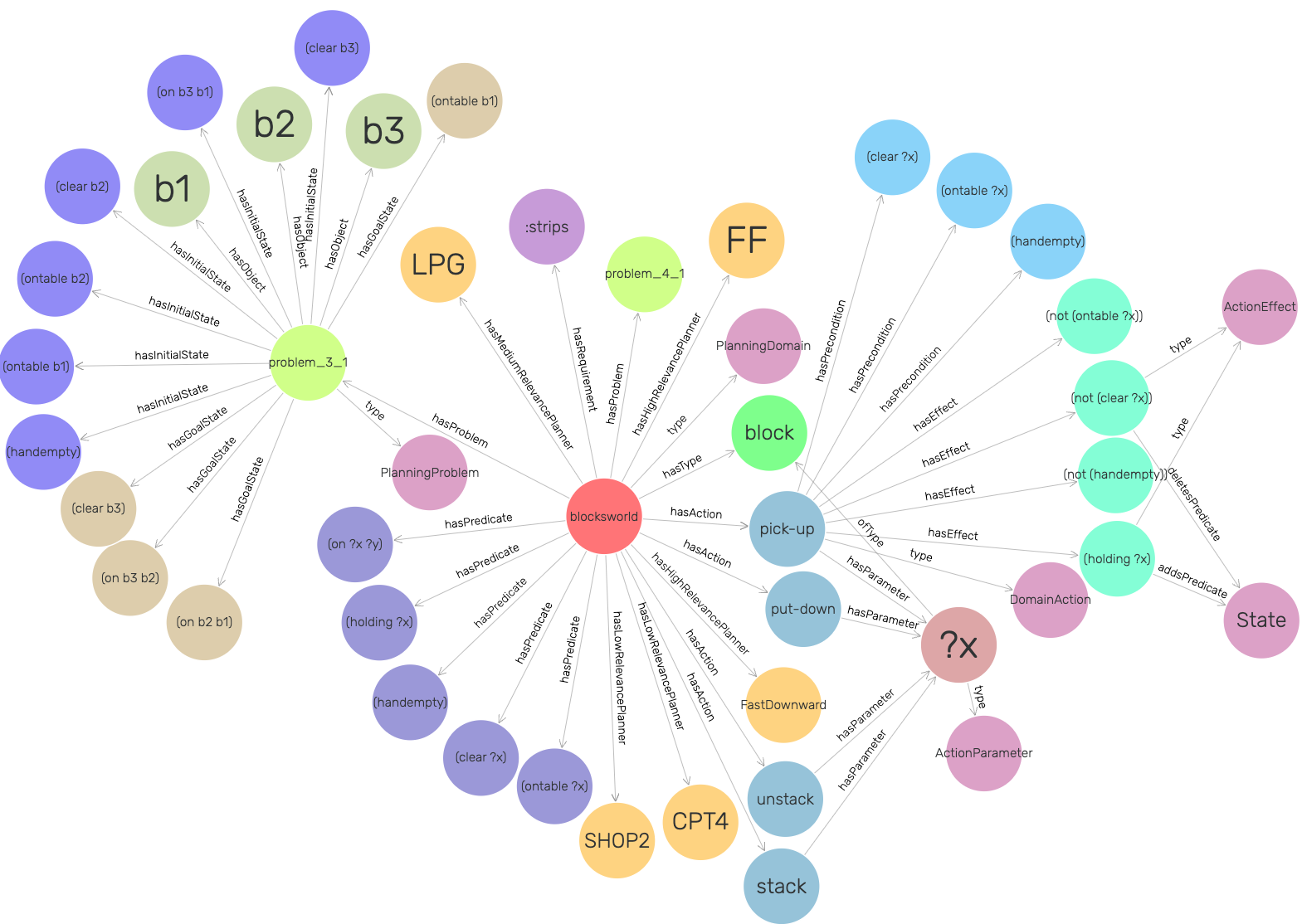}
    \caption{Knowledge graph representation for blocksworld domain from IPC-2000}
    \label{fig:bw_kg}
\end{figure}

\subsubsection{Evaluation of Competency questions:}
For the evaluation of the competency questions, we have considered a sample knowledge graph, shown in Figure \ref{fig:bw_kg}, for \verb|blocksworld| from IPC-2000 domain created using planning ontology shown in Figure \ref{fig:ontology}. SPARQL queries for each of these questions can be found at our GitHub Repository\textsuperscript{\ref{footnote: repo}}.

\begin{table}[!t]
\centering
\caption{Demonstrating the effectiveness of two different policies employed to choose a planner for problem-solving.}
\begin{tabular}{lcccc}
\hline
\multicolumn{1}{c}{\textbf{Domain}} & \multicolumn{2}{c}{\textbf{Ontology Policy}} & \multicolumn{2}{c}{\textbf{Random Policy}}  \\ \cline{2-5} 
\multicolumn{1}{c}{}                                 & \multicolumn{1}{c}{Avg. Exp.}  & Avg. Plan Cost & \multicolumn{1}{c}{Avg. Exp} & Avg. Plan Cost \\ \hline
scanalyzer                                             & \multicolumn{1}{c}{\textbf{8,588}}    & 20                                 & \multicolumn{1}{c}{8,706}    & 20                                 \\ 
elevators                                              & \multicolumn{1}{c}{\textbf{1,471}}    & 52                                 & \multicolumn{1}{c}{64,541}   & 52                                 \\ 
transport                                              & \multicolumn{1}{c}{165,263}           & 491                                & \multicolumn{1}{c}{\textbf{132,367}}  & 491                                \\ 
parking*                                               & \multicolumn{1}{c}{\textbf{367,910}}  & 18                                 & \multicolumn{1}{c}{488,830}  & 17                                 \\ 
woodworking                                            & \multicolumn{1}{c}{\textbf{1,988}}    & 211                                & \multicolumn{1}{c}{19,844}   & 211                                \\ 
floortile**                                            & \multicolumn{1}{c}{283,724}           & 54                                 & \multicolumn{1}{c}{\textbf{2,101}}    & 49                                 \\ 
barman                                                 & \multicolumn{1}{c}{\textbf{1,275,078}} & 90                                 & \multicolumn{1}{c}{5,816,476} & 90                                 \\ 
openstacks                                             & \multicolumn{1}{c}{\textbf{132,956}}  & 4                                  & \multicolumn{1}{c}{139,857}  & 4                                  \\ 
nomystery                                              & \multicolumn{1}{c}{1,690}             & 13                                 & \multicolumn{1}{c}{1,690}    & 13                                 \\ 
pegsol                                                 & \multicolumn{1}{c}{\textbf{89,246}}   & 6                                  & \multicolumn{1}{c}{101,491}  & 6                                  \\ 
visitall                                               & \multicolumn{1}{c}{5}                & 4                                  & \multicolumn{1}{c}{5}       & 4                                  \\ 
tidybot**                                              & \multicolumn{1}{c}{\textbf{1,173}}    & 17                                 & \multicolumn{1}{c}{3,371}    & 33                                 \\ 
parcprinter                                            & \multicolumn{1}{c}{541}              & 441,374                             & \multicolumn{1}{c}{\textbf{417}}     & 441,374                             \\ 
sokoban                                                & \multicolumn{1}{c}{\textbf{9,653}}    & 25                                 & \multicolumn{1}{c}{156,600}  & 25                                 \\ \hline
\end{tabular}
\label{tab:best_planner_eval}
\end{table}

\subsubsection{Usecase 1: Identifying Most Promising Planner} - 
One of the major challenges in the field of artificial intelligence (AI) is the automated selection of the best-performing planner for a given planning domain. This challenge arises due to the vast number of available planners and the diversity of planning domains. The traditional way to select a planner is to experiment with various search algorithms and heuristics and settle on an appropriate combination as seen in IPC competitions. To address this challenge, we now present a new approach by using our planning ontology to represent the features of the planning domain and the capabilities of planners.

The ontology for planning aims to capture the connection between the Planning Domain and the Planner by indicating the relevance of a planner to a specific domain. We made use of data acquired from International Planning Competitions (IPCs) to furnish specific details regarding the relevance of planners. The IPC results provide us with relevant details on the planners that took part in the competition and the domains that were evaluated during that particular year. This information includes specifics on how each planner performed against all the domains that participated.

To show the usage of extracting the most promising planners for a given domain, we have used IPC-2011 data\footnote[3]{\label{ipc-2011}http://www.plg.inf.uc3m.es/ipc2011-deterministic/} (optimal track). The ontology was populated with data acquired from the IPC-2011, which provided relevant details on the planners that took part in the competition and the domains that were evaluated during IPC-2011. A relevance relation of either \textit{low}, \textit{medium}, or \textit{high} was assigned to each planner based on the percentage, \textit{low-}below 35\%, \textit{medium-}35\% to 70\%, \textit{high-}70\% and above, of problems they solved in a given domain. In this experiment, we consider that the experimental environment has four planners available: Fast Downward Stone Soup 1\footnote[4]{\label{fd_ipcPlanners}https://www.fast-downward.org/IpcPlanners}, LM-Cut\textsuperscript{\ref{fd_ipcPlanners}}, Merge and Shrink\textsuperscript{\ref{fd_ipcPlanners}}, and BJOLP\textsuperscript{\ref{fd_ipcPlanners}}. We evaluate 3 problem instances of each domain from IPC-2011 with 2 policies for selecting planners to generate plans for each of these problem instances - 
\begin{enumerate}
    \item \textbf{Random Policy:} To solve each problem instance, this policy selects a random planner from the available planners.
    \item \textbf{Ontology Policy:} To solve each problem instance, this policy extracts the information on the best planner for the problem domain from the ontology populated with IPC-2011 data.
\end{enumerate}

Table \ref{tab:best_planner_eval} presents the results of our evaluation, indicating the average number of nodes expanded and plan cost for each policy in a given domain.
The table provides a comprehensive summary of the performance of different planners in terms of their efficiency and effectiveness. An ideal planner is expected to generate a solution with low values for both these metrics. 
The {\em Ontology Policy}, designed to select the best-performing planner for a given domain, outperformed the {\em Random Policy} in terms of the average number of nodes expanded to find a solution. Moreover, the {\em Random Policy} failed to solve problems in the parking (1 out of 3), floortile (2 out of 3), and tidybot (2 out of 3) domains, which highlights the limitations of choosing a planner randomly. But if a domain is easily solvable by relevant planners that can tackle them, {\em Random Policy} may still do well. 

What we demonstrate is a rather simple usage of the Ontology for Planner Selection policy. Creating more advanced strategies is a promising area for further research.




\subsubsection{Usecase 2: Extracting Macro Operators} -
While automated planning has been successful in many domains, it can be computationally expensive, especially for complex problems. One approach to improve efficiency is by using macro-operators, which are sequences of primitive actions that can be executed as a single step. However, identifying useful macro-operators manually can be time-consuming and challenging. Authors in \cite{chrpa2010generation} introduce a novel method for improving the efficiency of planners by generating macro-operators. The proposed approach involves analyzing the inter-dependencies between actions in plans and extracting macro-operators that can replace primitive actions without losing the completeness of the problem domain. The soundness and complexity of the method are assessed and compared to other existing techniques. The paper asserts that the generated macro-operators are valuable and can be seamlessly integrated into planning domains without losing the completeness of the problem. In \cite{botea2005learning}, the authors detail a three-step method for learning and utilizing macro-operators to enhance planning efficiency in new problems. Initially, a comprehensive set of macros is generated from the solution graphs of various training problems. This set is then narrowed down through a filtering process. The selected macros are subsequently applied to expedite problem-solving. The generation phase involves extracting and selecting specific subgraphs from solution graphs to create individual macros.

Based on the ontology depicted in Figure \ref{fig:ontology}, we extract macro-operators that can enhance the efficiency of planners. To demonstrate this, we have considered three different domains: \verb|blocksworld|(bw), \verb|driverlog|(dl), and \verb|grippers|(gr), presented in IPC-2000, 2002, and 1998 respectively. We initially developed a knowledge graph using the ontology represented in Figure \ref{fig:ontology} for the three domains of interest. Subsequently, we employed a SPARQL query to retrieve the stored plans for these domains. We then examined these plans to identify the sequences of action pairs and ranked them based on their frequency of occurrence. To improve the effectiveness of this technique, it is essential to consider both the frequency of occurrence of action pairs and the properties of the domain. Specifically, the precondition and effect of actions should be analyzed to ensure that the first action leads to the precondition of the second action in the pair. We employed another SPARQL query to extract the preconditions and effects associated with each of these actions. We analyzed the resulting action pairs to verify their validity of occurrence, thereby filtering out pairs that did not have a combined effect. The results of this extraction process are shown in Table \ref{tab: macros}. These action relations are stored back into the knowledge graph in the \verb|MacroAction| class and can be utilized by planners to enhance their efficiency.


\begin{table}[t]
\centering
\caption{Extracted action relations, ordered based on their frequency, for domains \texttt{blocksworld}, \texttt{driverlog}, and \texttt{grippers}.}
\begin{tabular}{p{2.5cm}p{9cm}}
\hline
\textbf{Domains} & \textbf{Extracted Action Relations} \\
\hline
\cellcolor{blue!25}\texttt{blocksworld} & \texttt{unstack} * \texttt{put-down}; \texttt{pick-up} * \texttt{stack}; \texttt{put-down} * \texttt{unstack}; \texttt{stack} * \texttt{pick-up}; \texttt{unstack} * \texttt{stack}; \texttt{put-down} * \texttt{pick-up}; \texttt{stack} * \texttt{unstack}\\
\hline
\cellcolor{green!25}\texttt{driverlog} & \texttt{drive-truck} * \texttt{unload-truck}; \texttt{drive-truck} * \texttt{load-truck}; \texttt{board-truck} * \texttt{drive-truck}; \texttt{walk} * \texttt{board-truck}\\
\hline
\cellcolor{red!25}\texttt{grippers} & \texttt{pick} * \texttt{move}; \texttt{move} * \texttt{drop}\\
\hline
\end{tabular}

\label{tab: macros}
\end{table}

\begin{table}[b]
\centering
\begin{tabular}{lcccccc}
\hline
 \multicolumn{1}{c}{\textbf{Domain}} & \multicolumn{3}{c}{\textbf{Original Domain}} & \multicolumn{3}{c}{\textbf{Domain With Macros}} \\ \cline{2-7}
\multicolumn{1}{c}{} & \multicolumn{1}{c}{Avg. Exp.} & \multicolumn{1}{c}{Avg. Eval.} & Avg. Gen. & \multicolumn{1}{c}{Avg. Exp.} & \multicolumn{1}{c}{Avg. Eval.} & Avg. Gen. \\ \hline
\textbf{blocksworld} & \multicolumn{1}{c}{20219}      & \multicolumn{1}{c}{59090}      & 106321     & \multicolumn{1}{c}{18}        & \multicolumn{1}{c}{310}        & 359       \\
\textbf{gripper} & \multicolumn{1}{c}{2672}      & \multicolumn{1}{c}{10660}      & 30871     & \multicolumn{1}{c}{510}        & \multicolumn{1}{c}{3974}        & 11468       \\
\textbf{driverlog} & \multicolumn{1}{c}{3753}      & \multicolumn{1}{c}{17849}      & 45753     & \multicolumn{1}{c}{14888}        & \multicolumn{1}{c}{720008}        & 209760       \\ \hline

\end{tabular}

\caption{Comparison of planner performance between original and macro-enabled versions of three planning domains, showing the average number of nodes expanded, evaluated, and generated.}

\label{tab: macros_results}
\end{table}

Table \ref{tab: macros_results} shows the comparison of a planner performance given the original domain and macros-enabled version of the domain. For this evaluation, we have considered the FastDownward planner \cite{helmert2006fast} with LM-Cut Heuristic \cite{helmert2011lm} to generate plans for 20 problems of varying complexities for each domain. We evaluate the performance of each domain based on the average number of nodes expanded, evaluated, and generated to find a solution. This study demonstrates that macro operators can enhance the planner performance in most of the domains tested, with the exception of the \verb|driverlog| domain. In this domain, the planner performs worse when macro operators are included, as they increase the average number of nodes expanded, evaluated, and generated. This is due to the fact that the macro operators introduce more actions to the domain, which increases the branching factor and challenges the heuristic to select the optimal action at each step. Hence, the applicability of macro operators depends on the features of the domain and the planner. Macro operators can facilitate the planning process by decreasing the search depth, but they can also hinder it by increasing the search width. A potential improvement is to use a more informative heuristic that guides the planner to choose the best action at each step.

\section{Conclusion}

In this work, we build and share a planning ontology that provides a structured representation of concepts and relations for planning, allowing for efficient extraction of domain, problem, and planner properties. The ontology's practical utility is demonstrated in identifying the best-performing planner for a given domain and extracting macro operators using plan statistics and domain properties. Standardized benchmarks from IPC domains and planners offer an objective and consistent approach to evaluating planner performance, enabling rigorous comparisons in different domains to identify the most suitable planner. The planning ontology can aid researchers and practitioners in automated planning, and its use can simplify planning tasks and boost efficiency. As the field of AI planning continues to evolve, planning ontology can play a crucial role in advancing the state-of-the-art while leveraging the past.


Future work could explore the use of a mixed reasoning strategy that combines the structured, top-down approach of ontologies with the dynamic, bottom-up capabilities of Large Language Models (LLMs) \cite{Mittal2017ThinkingFA-ontology}. This approach can be particularly effective in contexts like LLMs, which have shown promise for automated planning \cite{plansformer-paper-pallagani2022}. Furthermore, our ontology, with its specific data properties for storing Action explanations, can be leveraged to enhance this hybrid model. It can provide comprehensive explanations for planning decisions as shown in the workflow Figure \ref{fig:use}, adding an interpretive layer that is crucial for complex domains such as multi-agent systems, where understanding the rationale behind each agent's actions is key. This blend of ontology-based clarity and LLM-driven adaptability could offer nuanced insights into coordinating actions and explaining them in a way that is both transparent and informative.

\bibliographystyle{splncs04}
%
\bibliography{references}




\end{document}